\documentclass[runningheads]{llncs}
\usepackage{graphicx}
\usepackage{comment}
\usepackage{amsmath,amssymb} 
\usepackage{color}
\usepackage[svgnames]{xcolor}
\usepackage{epsfig}
\usepackage{amsmath}
\usepackage{bm}
\usepackage{amssymb}
\usepackage{algorithm}
\usepackage{hyperref}
\usepackage{enumitem}
\usepackage{breakcites}
\hypersetup{
    colorlinks=true,
    linkcolor=blue,
    filecolor=magenta,      
    urlcolor=blue,
}
\usepackage{lipsum}
\usepackage{url}

\begin{document}
\pagestyle{headings}
\mainmatter
\def\ECCVSubNumber{2116}  

\title{Symbiotic Adversarial Learning for Attribute-based Person Search} 

\titlerunning{Symbiotic Adversarial Learning for Attribute-based Person Search}
%
\author{Yu-Tong Cao$^*$  \and
Jingya Wang$^*$  \and
Dacheng Tao}
\authorrunning{Cao Y.-T, Wang J., and Tao D.}
%
\institute{UBTECH Sydney AI Centre, School of Computer Science, Faculty of Engineering, The University of Sydney, Darlington, NSW 2008, Australia \\
\email{\{ycao5602@uni.,jingya.wang@,dacheng.tao@\}sydney.edu.au}}
\maketitle

\newcommand\blfootnote[1]{%
\begingroup
\renewcommand\thefootnote{}\footnote{#1}%
\addtocounter{footnote}{-1}%
\endgroup
}

\begin{abstract}
Attribute-based person search is in significant demand for applications where no detected query images are available, such as identifying a criminal from witness. However, the task itself is quite challenging because there is a huge modality gap between images and physical descriptions of attributes. Often, there may also be a large number of unseen categories (attribute combinations). 
The current state-of-the-art methods either focus on learning better cross-modal embeddings by mining only seen data, or they explicitly use generative adversarial networks (GANs) to synthesize unseen features. The former tends to produce poor embeddings due to insufficient data, while the latter does not preserve intra-class compactness during generation. 
In this paper, we present a symbiotic adversarial learning framework, called SAL. Two GANs sit at the base of the framework in a symbiotic learning scheme: one synthesizes features of unseen classes/categories, while the other optimizes the embedding and performs the cross-modal alignment on the common embedding space. 
Specifically, two different types of generative adversarial networks learn collaboratively throughout the training process and the interactions between the two mutually benefit each other.
Extensive evaluations show SAL’s superiority over nine state-of-the-art methods with two challenging pedestrian benchmarks, PETA and Market-1501. The code is publicly available at: \url{https://github.com/ycao5602/SAL}. 

\end{abstract}

\section{Introduction}
%
The goal with “person search” is to find the same person in non-overlapping camera views at different locations. In surveillance analysis, it is a crucial tool for public safety.
To date, the most common approach to person search has been to take one detected image captured from a surveillance camera and use it as a query \cite{li2014deepreid,wang2016joint,zhang2016learning,hermans2017defense,zhong2017re,sun2017svdnet,li2017person,chen2018person}. However, this is not realistic in many real-world applications – for example, where a human witness has identified the criminal, but no image is available.\blfootnote{$^*$ Equal contribution.}


Attributes, such as gender, age, clothing or accessories, are more natural to us as searchable descriptions, and these can be used as soft biometric traits to search for in surveillance data \cite{liu2012person,layne2012person,jaha2014soft,reid2014soft}. 
Compared to the queries used in image-based person search, these attributes are also much easier to obtain.  Further, semantic descriptions are more robust than low-level visual representations, where changes in viewpoint or diverse camera conditions can be problematic. Recently, a few studies have explored sentence-based person search \cite{li2017person,li2017identity}.
Although this approach provides rich descriptions, unstructured text tends to introduce noise and redundancy during modeling. Attributes descriptions, on the other hand, are much cheaper to collect, and they inherently have a robust and relatively independent ability to discriminate between persons. 
As such, attribute descriptions have the advantage of efficiency over sentence descriptions in person search tasks.

Unfortunately, applying a cross-modal, attribute-based person search to real-world surveillance images is very challenging for several reasons.  
(1) There is a huge semantic gap between visual and textual modalities (i.e., attribute descriptions). Attribute descriptions are of lower dimensionality than visual data, e.g., tens versus thousands, and they are very sparse. Hence, in terms of data capacity, they are largely inferior to visual data. From performance comparisons between single-modal retrieval (image-based person search) and cross-modal retrieval (attribute-based person search) using current state-of-the-art methods, there is still a significant gap between the two, e.g., mAP $84.0\%$~\cite{saquib2018pose}
vs. $24.3\%$~\cite{dong2019person} on Market-1501 ~\cite{zheng2015scalable,lin2017improving}.
%
(2) Most of the training and testing classes (attribute combinations) are non-overlapping, which leads to zero-shot retrieval problems – a very challenging scenario to deal with \cite{dong2019person}. Given an attribute-style query, model aims to have more capacity to search an unseen class given a query of attributes. 
%
(3) Compared to the general zero-shot learning settings one might see in classification tasks, surveillance data typically has large intra-class variations and inter-class similarities with only a small number of available samples per class (Fig.~\ref{fig:dataset}), e.g., $\sim 7$ samples per class in PETA~\cite{deng2014pedestrian} and $\sim 26$ samples per class in Market-1501 compared with $\sim 609$ samples per class in AWA2~\cite{xian2018zero}. One category (i.e., general attribute combinations that are not linked to a specific person) may include huge variations in appearance – just consider how many dark-haired, brown-eyed people you know and how different each looks. Also, the inter-class distance between visual representations from different categories can be quite small considering fine-grained attributes typically become ambiguous in low resolution with motion blur. 
%

\begin{figure}[t]
  \centering
  \includegraphics[trim={0 11cm 0 0},clip,page=2,width=\textwidth]{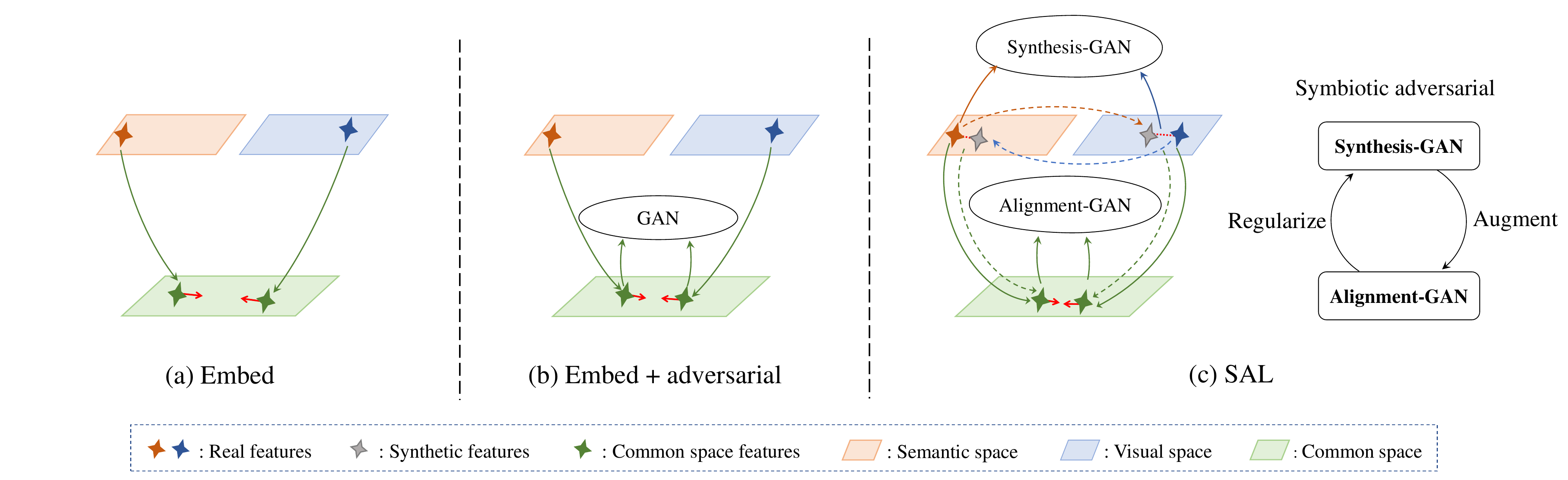}
  \caption{\footnotesize Three categories of cross-modal retrieval with common space learning: (a) The embedding model that projects data from different modalities into common feature space. (b) The embedding model with common space adversarial alignment. (c) The proposed SAL that jointly considers feature-level data augmentation and cross-modal alignment by two GANs.}
  \label{fig:retrieval}
\end{figure}
One general idea for addressing this problem is to use cross-modal matching to discover a common embedding space for the two modalities. Most existing methods focus on representation learning, where the goal is to find projections of data from different modalities and combine them in a common feature space. This way, the similarity between the two can be calculated directly \cite{hardoon2004canonical,wang2013learning,feng2014cross} (Fig.~\ref{fig:retrieval}(a)). However, these approaches typically have a weaker modality-invariant learning ability and, therefore, weaker performance in cross-modal retrieval. More recently, some progress has been made with the development of GANs \cite{goodfellow2014generative}. GANs can better align the distributions of representations across modalities in a common space \cite{wang2017adversarial}, plus they have been successfully applied to attribute-based person search \cite{yin2018adversarial}. (Fig.~\ref{fig:retrieval}(b)). There are still some bridges to cross, however. With only a few samples, only applying cross-modal alignment to a high-level common space may mean the model fails to capture variances in the feature space. Plus, the cross-modal alignment probably will not work for unseen classes. Surveillance data has all these characteristics, so all of these problems must be overcome. 

To deal with unseen classes, some recent studies on zero-shot learning have explicitly used GAN-based models to synthesize those classes ~\cite{long2017zero,kumar2018generalized,zhu2018generative,felix2018multi}. Compared to zero-shot classification problems, our task is cross-modal retrieval, which requires learning a more complex search space with a finer granularity. As our experiments taught us, direct generation without any common space condition may reduce the intra-class compactness.

\begin{figure}[t]
  \centering
  \includegraphics[page=2,width=0.8\textwidth]{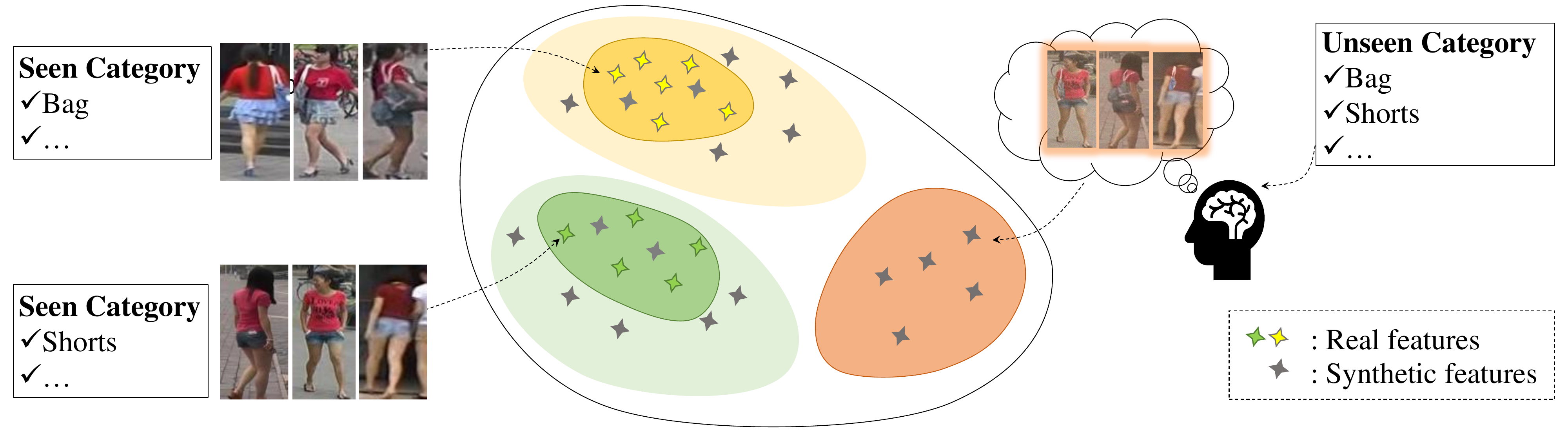}
  \caption{\footnotesize An illustration of feature space data augmentation with a synthesized unseen class.}
  \label{fig:augmentation}
\end{figure}

Hence, in this work, we present a fully-integrated symbiotic adversarial learning framework for cross-modal person search, called SAL. Inspired by symbiotic evolution, where two different organisms cooperate so both can specialize, we jointly explore the feature space synthesis and common space alignment with separate GANs in an integrated framework at different scales (Fig.~\ref{fig:retrieval}(c)). The proposed SAL mainly consists of two GANs with interaction: (1) A synthesis-GAN that generates synthetic features from semantic representations to visual representations, and vice versa. The features are conditioned on common embedding information so as to preserve very fine levels of granularity. 
(2) An alignment-GAN that optimizes the embeddings and performs cross-modal alignment on the common embedding space. In this way, one GAN augments the data with synthetic middle-level features, while the other uses those features to optimize the embedding framework. Meanwhile, a new regularization term, called common space granularity-consistency loss, forces the cross-modal generated representations to be consistent with their original representations in the high-level common space. These more reliable embeddings further boost the quality of the synthetic representations. 
To address zero-shot learning problems when there is no visual training data for an unseen category, we use new categories of combined attributes to synthesize visual representations for augmentation. 
An illustration of feature space data augmentation with a synthesized unseen class is shown in Fig.~\ref{fig:augmentation}.

In summary, this paper makes the following main {\bf contributions} to the literature:
\begin{itemize}
\item  We propose a fully-integrated framework called symbiotic adversarial learning (SAL) for attribute-based person search. The framework jointly considers feature-level data augmentation and cross-modal alignment by two GANs at different scales. Plus, it handles cross-model matching, few-shot learning, and zero-shot learning problems in one unified approach.
\item We introduce a symbiotic learning scheme where two different types of GANs learn collaboratively and specially throughout the training process. By generating qualified data to optimise the embeddings, SAL can better align the distributions in the common embedding space, which, in turn, yields superior cross-modal search results.
\item Extensive evaluations on the PETA~\cite{deng2014pedestrian} and Market-1501~\cite{zheng2015scalable} datasets demonstrate the superiority of SAL at attribute-based person search over nine state-of-the-art attribute-based models.
\end{itemize}  
       
\section{Related Works}

 \noindent {\bf Attribute-Based Person Search.}
In recent years, semantic attribute recognition of pedestrians has drawn increasing attention \cite{li2015multi,layne2014attributes,deng2015learning,wang2017attribute,zhao2018grouping} and it has been extensively exploited for image-based person search as a middle-level feature \cite{layne2012towards,su2015multi,su2016deep,lin2017improving,wang2018transferable}. Some studies on exploit attribute-based person search for cross-modal retrieval \cite{vaquero2009attribute,siddiquie2011image,scheirer2012multi,yin2018adversarial}.
Early attribute-based retrieval methods intuitively rely on attribute prediction.
For example, Vaquero et al. \cite{vaquero2009attribute} proposed a human parts based attribute recognition method for person retrieval. Siddiquie et al. \cite{siddiquie2011image} utilized a structured prediction framework to integrate ranking and retrieval with complex queries, while Scheirer et al. \cite{scheirer2012multi} constructed normalized “multi-attribute spaces” from raw classifier outputs.
However, the pedestrian attribute recognition problem is far from being solved \cite{li2015multi,wang2017attribute,liu2017hydraplus}. Consistently predicting all the attributes is a difficult task with sparsely-labeled training data: the images of people from surveillance cameras are low resolution and often contain motion blur. Also the same pedestrian would rarely be captured in the same pose. These “imperfect attributes” are significantly reducing the reliability of existing models. Shi et al. \cite{shi2015transferring} suggested transfer learning as a way to overcome the limited availability of costly labeled surveillance data. They chose fashion images with richly labeled captions as the source domain to bridge the gap between unlabeled pedestrian surveillance images. They were able to produce semantic representations for a person search without annotated surveillance data, but the retrieval results were not as good as the supervised/semi-supervised methods, which raises a question of cost versus performance.  \cite{yin2018adversarial} posed attribute-based person search as a cross-modality matching problem. They applied an adversarial leaning method to align the cross-modal distributions in a common space. However, their model design did not extend to the unseen class problem or few-shot learning challenges. Dong et al. \cite{dong2019person} formulated a hierarchical matching model that fuses the similarity between global category-level embeddings and local attribute-level embeddings. Although they consider a zero-shot learning paradigm in the approach, the main disadvantage of this method is that it lacks the ability to synthesize visual representations of an unseen category. Therefore, it does not successfully handle unseen class problem for cross-modal matching.

\noindent {\bf Cross-Modal Retrieval.}
Our task of searching for people using attribute descriptions is closely related to studies on cross-modal retrieval as both problems require the attribute descriptions to be aligned with image data. This is a particularly relevant task to text-image embedding \cite{yan2015deep,wang2016learning,li2017person,wang2017adversarial,zheng2017dual,tsai2017learning}, where canonical correlation analysis (CCA) \cite{hardoon2004canonical} is a common solution for aligning two modalities by maximizing their correlation. With the rapid development of deep neural networks (DNN), a deep canonical correlation analysis (DCCA) model has since been proposed based on the same insight \cite{andrew2013deep}. Yan et al. \cite{yan2015deep} have subsequently extended the idea to an image-text matching method based on DCCA. Beyond these core works, a variety of cross-modal retrieval methods have been proposed for different ways of learning a common space. Most use category-level (categories) labels to learn common representations \cite{yin2018adversarial,wang2017adversarial,zhu2018generative,zheng2017dual,zheng2017dual}.
However, what might be fine-grained attribute representations often lose granularity, and semantic categories are all treated as being the same distance apart. Several more recent works are based on using a GAN \cite{goodfellow2014generative} for cross-modal alignment in the common subspace \cite{wang2017adversarial,yin2018adversarial,zhu2018generative,tzeng2017adversarial}. The idea is intuitive since GANs 
have been proved to be powerful tools for distribution alignment \cite{tzeng2017adversarial,hoffman2017cycada}. 
Further, they have produced some impressive results in image translation \cite{zhu2017unpaired}, domain adaptation \cite{tzeng2017adversarial,hoffman2017cycada} and so on. However, when only applied to a common space, conventional adversarial learning may fail to model data variance where there are only a few samples and, again, the model design ignores the zero-shot learning challenge.

\section{Symbiotic Adversarial Learning (SAL) }

\noindent {\bf Problem Definition }
Our deep model for attribute-based person search is based on the following specifications. There are $N$ labeled training images $\{\bm{x}_i,\bm{a}_i,y_i\}_{i=1}^{N}$
available in the training set. 
Each image-level attribute description $\bm{a}_i = [a_{(i,1)}, \dots, a_{(i,n_\text{attr})}]$ 
is a binary vector with $n_\text{attr}$ number of
attributes, where $0$ and $1$ indicate the absence and presence 
of the corresponding attribute.
Images of people with the same attribute description are assigned to a unique category so as to derive a category-level label, specifically $y_i\in \{1,...,M\}$ for $M$ categories.
The aim is to find matching pedestrian images from the gallery set $\mathcal{X}_\text{gallery} = \{\bm{x}_j\}_{j=1}^{G}$ with $G$ images given an attribute description $\bm{a}_q$ from the query set. 

\subsection{Multi-modal Common Space Embedding Base }
\label{sec:Embedding}

One general solution for cross-modal retrieval problems is to learn a common joint subspace where samples of different modalities can be directly compared to each other. As mentioned in our literature review, most current approaches use category-level labels (categories) to generate common representations \cite{yin2018adversarial,wang2017adversarial,zhu2018generative}. However, these representations never manage to preserve the full granularity of finely-nuanced attributes. Plus, all semantic categories are treated as being the same distance apart when, in reality, the distances between different semantic categories can vary considerably depending on the similarity of their attribute descriptions.

Thus, we propose a common space learning method that jointly considers the global category and the local fine-grained attributes. The common space embedding loss is defined as:

\begin{equation}
L_\text{embed}=L_\text{cat} + L_\text{att}.
\label{eq:concept}
\end{equation}
The category loss utilizing the Softmax Cross-Entropy loss function for image embedding branch is defined as:

\begin{equation}
L_\text{cat}=-\sum_{i=1}^{N}\log \Big( p_\text{cat}(\bm{x}_{i}, y_i) \Big),
\label{eq:CE_softmax}
\end{equation}
where $p_\text{cat}(\bm{x}_{i}, y_i)$ specifies the predicted probability on the ground truth 
category $y_i$ of the training image $\bm{x}_{i}$.

To make the common space more discriminative for fine-trained attributes,
we use the Sigmoid Cross-Entropy loss function which considers all $m$ attribute classes:
\begin{align}\label{eq:CE-sigmoid}
\small
L_\text{att}=-\sum_{i=1}^{N}
\sum_{j=1}^{m}
\Big( 
a_{(i,j)} \log\big( p^{(j)}_\text{att}(\bm{x}_i) \big) + 
(1-a_{(i,j)}) \log \big( 1 - p^{(j)}_\text{att}(\bm{x}_i) \big)
\Big),
\end{align}
where $a_{(i,j)}$ and $p^{(j)}_\text{att}(\bm{x}_i)$ 
define the ground truth label
and the predicted classification probability on the $j$-th attribute class
of the training image $\bm{x}_i$,
i.e. $\bm{a}_i = [a_{(i,1)},\cdots,a_{(i,m)}]$
and $\bm{p}^{(j)}_{\text{att}} = [p^{(1)}_\text{att}(\bm{x}_i),\cdots,p^{(m)}_\text{att}(\bm{x}_i)]$.
Similar loss functions of Eq.~\eqref{eq:CE_softmax} and Eq.~\eqref{eq:CE-sigmoid} are applied to the attribute embedding branch as well. 

The multi-modal common space embedding can be seen as our base, named as \textit{Embed} (Fig.~\ref{fig:retrieval}(a)). A detailed comparison is shown in Table~\ref{tab:ablation}.

\begin{figure}[t]
  \centering
  \includegraphics[width=\textwidth, page=1]{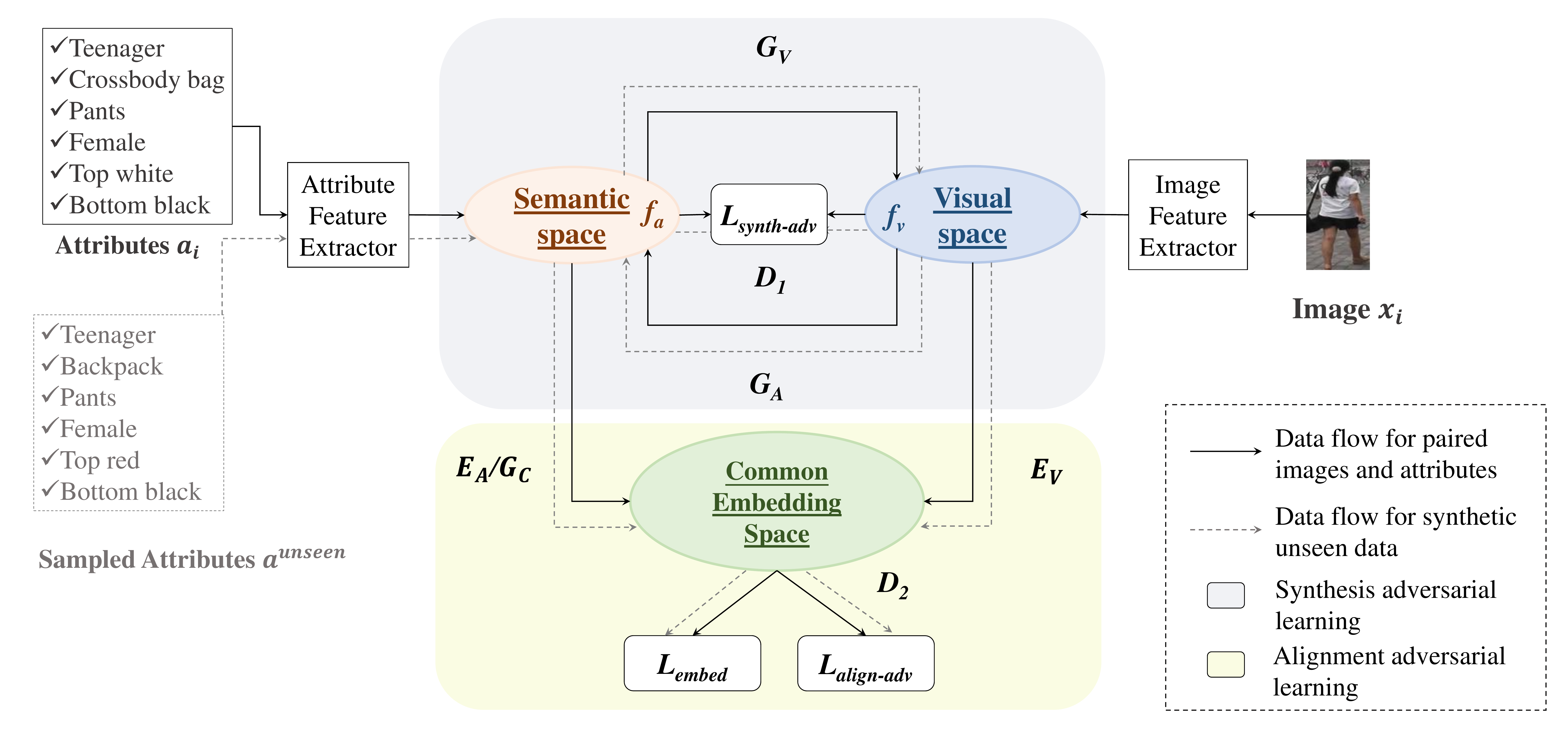}
  \caption{\footnotesize An overview of the proposed SAL Framework. 
  }
  \label{fig:full-model}
\end{figure}


\subsection{Middle-Level Granularity-Consistent Cycle Generation }
\label{sec:middle}
To address the challenge of insufficient data and bridge the modality gap between middle-level visual and semantic representations, we aim to generate synthetic features from the semantic representations to supplement the visual representations and vice versa. The synthetic features are conditioned on common embedding information so as to preserve very fine levels of granularity for retrieval. The middle-level representations are denoted as $f_v$ for visual and $f_a$ for attribute. More detail is provided in Fig.~\ref{fig:full-model}. 

As paired supervision are available for seen categories during generation, conventional unsupervised cross-domain generation methods, e.g. CycleGAN~\cite{zhu2017unpaired}, DiscoGAN~\cite{kim2017learning} are not perfect suit. To better match joint distributions, we consider single discriminator distinguishes whether the paired data $(f_{a},f_{v})$ ) is from a real feature distribution $ p (f_{a},f_{v})$ or not. This is inspired by  Triple Generative Adversarial Networks \cite{chongxuan2017triple}. The TripleGAN works on semi-supervised classification task with a generator, a discriminator and a classifier while our synthesis-GAN consists of two generators and a discriminator. 
As shown in Fig.~\ref{fig:full-model}, our synthesis-GAN consists of three main components: Generator $G_{A}$ synthesizes semantic representations from visual representations; Generator $G_{V}$ synthesizes visual representation from semantic representations, and a discriminator $D_1$ focused on identifying synthetic data pairs. Given that semantic representations are rather more sparse than visual representations, we add a noise vector $z\sim \mathcal{N}(0,1)$ sampled from a Gaussian distribution for $G_{V}$ to get variation in visual feature generation. Thus, $G_{V}$ can be regarded as a one-to-many matching from a sparse semantic space to a rich visual space. This accords with the one-to-many relationship between attributes and images. The training process is formulated as a minmax game between three players - $G_{V}$, $G_{A}$ and $D_\text{1}$ - where, from a game theory perspective, both generators can achieve their optima. The adversarial training scheme for middle-level cross-modal feature generation can, therefore, be formulated as:

\begin{flalign}
\label{eq:gan-m}
\begin{split} 
&L_\text{gan1}(G_{A},G_{V},D_\text{1})=\mathbb{E}_{(f_{a},f_{v})\sim p(f_{a},f_{v}))}[log(D_\text{1}(f_{a},f_{v}))]\\ 
&+ \frac{1}{2}\mathbb{E}_{f_{a}\sim p(f_{a})}[log(1-D_\text{1}(f_{a},\widetilde{f_{v}}))]
  + \frac{1}{2}\mathbb{E}_{f_{v}\sim{p(f_{v})}}[log(1-D_\text{1}(\widetilde{f_{a}},f_{v}))].
\end{split}
\end{flalign}
The discriminator $D_\text{1}$ is fed with three types of input pairs: (1) The fake input pairs $(\widetilde{f_{a}},f_{v})$, where $\widetilde{f_{a}}= G_{A}(f_{v})$. (2) The fake input pairs $(f_{a},\widetilde{f_{v}})$, where $\widetilde{f_{v}}=G_{V}(f_{a},z))$. (3) The real input pairs $(f_{a},f_{v}) \sim p (f_{a},f_{v})$.

To constrain the generation process to only produce representations that are close to the real distribution, we assume the synthetic visual representation can be generated back to the original semantic representation, which is inspired by the CycleGAN structure~\cite{zhu2017unpaired}. 
Given this is a one-to-many matching problem as mentioned above, we flex the ordinary two-way cycle consistency loss to a one-way cycle consistency loss (from the semantic space to the visual space and back again): 

\begin{flalign}
\label{eq:cyc}
\begin{split} 
L_\text{cyc}(G_{A},G_{V}) = \mathbb{E}_{f_{a}\sim p(f_{a})}[||G_{A}(G_{V}(f_{a},z))-f_{a}||_{2}].
\end{split}
\end{flalign}

Furthermore, feature generation is conditioned on embeddings in the high-level common space, with the aim of generating more meaningful representations that preserve as much granularity as possible. This constraint is a new regularization term that we call “the common space granularity-consistency loss”, formulated as follows:

\begin{flalign}
\begin{split} 
&L_\text{consis}(G_{A},G_{V})= 
\mathbb{E}_{f_{v}\sim p(f_{v})}[||E_{A}(\widetilde{f_a})-E_{V}(f_{v})||_{2}] 
+ \mathbb{E}_{f_{a}\sim p(f_{a})}[||E_{V}(\widetilde{f_v})-E_{A}(f_{a})||_{2}]\\ 
&+  \mathbb{E}_{(f_{a},f_{v})\sim p(f_{a},f_{v})}[||E_{A}(\widetilde{f_a})-E_{A}(f_{a})||_{2}] 
 + \mathbb{E}_{(f_{a},f_{v})\sim p(f_{a},f_{v})}[||E_{V}(\widetilde{f_v})-E_{V}(f_{v})||_{2}],  \\
\end{split}
\label{eq:conc}
\end{flalign}
where $E_A$ and $E_V$ stand for the common space encoders for attribute features and visual features respectively.

Lastly, the full objective of synthesis-GAN is:

\begin{flalign}
\begin{split} 
L_\text{synth-adv}= L_\text{gan1} + L_\text{cyc} + L_\text{consis}.
\end{split}
\label{eq:adv1}
\end{flalign}



\subsection{High-Level Common Space Alignment with Augmented Adversarial Learning}
\label{sec:concept}
On the high-level common space which is optimized by Eq. \eqref{eq:concept}, we introduce the second adversarial loss for cross-modal alignment, making the common space more modality-invariant. Here, $E_A$ works as a synchronous common space encoder and generator. Thus, we can define $E_A$ as $G_C$ in this adversarial loss:

\begin{flalign}
\begin{split} 
L_\text{gan2}(G_{C},D_\text{2}) =& \mathbb{E}_{f_{v}\sim p(f_{v})}[logD_\text{2}(E_{V}(f_{v}))]\\ 
  +  &\mathbb{E}_{f_{a}\sim p(f_{a})}[log(1-D_\text{2}(E_{A}(f_{a})))].
\end{split}
\label{eq:gan2}
\end{flalign}
Benefit from the middle-level GAN that bridges the semantic and visual spaces, we can access the augmented data, $\widetilde{f_a}$ and $\widetilde{f_v}$, to optimise the common space, where $\widetilde{f_a} = G_{A}(f_{v})$ and $\widetilde{f_v} = G_{V}(f_{a},z)$. 
Thus the augmented adversarial loss is applied for cross-modal alignment by:

\begin{flalign}
\begin{split} 
L_\text{aug1}(G_{C},D_\text{2}) =& \mathbb{E}_{f_{a}\sim p(f_{a})}[logD_\text{2}(E_{V}(\widetilde{f_v}))]\\ 
   +  &\mathbb{E}_{f_{v}\sim p(f_{v})}[log(1-D_\text{2}(E_{A}(\widetilde{f_a}))].
\end{split}
\label{eq:agu1}
\end{flalign}
And can be used to optimize the common space by: 

\begin{flalign}
\begin{split} 
L_\text{aug2}(E_{A},E_{V}) = L_\text{embed}(\widetilde{f_a}) +L_\text{embed}(\widetilde{f_v}).
\end{split}
\label{eq:agu2}
\end{flalign}
The total augmented loss from the synthesis-GAN interaction is calculated by: 

\begin{flalign}
\begin{split} 
L_\text{aug} = L_\text{aug1} +L_\text{aug2}.
\end{split}
\label{eq:augtotal}
\end{flalign}
The final augmented adversarial loss for alignment-GAN is defined as: 

\begin{flalign}
\begin{split} 
L_\text{align-adv}= L_\text{gan2} + L_\text{aug}.
\end{split}
\label{eq:adv2}
\end{flalign}

\subsection{Symbiotic Training Scheme for SAL}
\label{sec:scheme}
Algorithm~\ref{Algorithm} summarizes the training process. As stated above, there are two GANs learn collaboratively and specially throughout the training process. The synthesis-GAN (including $G_{A}$, $G_{V}$ and $D_\text{1}$) aims to build a bridge of semantic and visual representations in the middle-level feature space, and to synthesize features from both to each other. The granularity-consistency loss was further proposed to constrain the generation that conditions on high-level embedding information. Thus, the high-level alignment-GAN (including $E_{A}$, $E_{V}$ and $D_\text{2}$) that optimizes the common embedding space benefits the synthesis-GAN with the better common space constrained by Eq.~\eqref{eq:conc}. Similarly, while the alignment-GAN attempts to optimize the embedding and perform the cross-modal alignment, the synthesis-GAN supports the alignment-GAN with more realistic and reliable augmented data pairs via Eqs.~\eqref{eq:agu1} and \eqref{eq:agu2}. Thus, the two GANs update iteratively with SAL’s full objective becomes:

\begin{flalign}
\begin{split} 
L_\text{SAL}=  L_\text{embed} + L_\text{synth-gan} + L_\text{align-gan}. 
\end{split}
\end{flalign}

\begin{algorithm}[t]
	\small
	\caption{\small Learning the SAL model.}
	\centering
	\label{Algorithm}
	\parbox{4.25in}{ 
		\textbf{Input:} $N$ labeled data pairs $(\bm{x}_i, \bm{a}_i)$ from the training set;
		\\ [0.05cm]
		\textbf{Output:} SAL attribute-based person search model; 
		\\[0.05cm]
		\textbf{for} $t=1$ \textbf{to}  \textsl{max-iteration} \textbf{do} \\ 
		\hphantom{~~~~~~} 
		Sampling a batch of paired training data $(\bm{x}_i, \bm{a}_i)$; \\
		\hphantom{~~~~~~} 
		\textbf{Step 1: Multi-modal common space Embedding}: \\
		\hphantom{~~~~~~} 
		Updating both image branch and attribute branch by common space embedding loss $L_\text{embed}$ (Eq. \eqref{eq:concept});\\
		\hphantom{~~~~~~} 
		\textbf{Step 2: Middle-Level Granularity-Consistent Cycle Generation }:
		Updating $G_{A}$, $G_{V}$ and $D_\text{1}$ by $L_\text{synth-adv}$ (Eq. \eqref{eq:adv1});\\
		\hphantom{~~~~~~} 
		\textbf{Step 3: High-Level Common Space Alignment}:
		Updating $E_{A}$, $E_{V}$ and $D_\text{2}$ by $L_\text{align-adv}$ (Eq. \eqref{eq:adv2});\\
		\textbf{end for} 
	}
\end{algorithm}
\noindent{\bf Semantic augmentation for unseen classes.} 
To address the zero-shot problem, our idea is to synthesize visual representations of unseen classes from semantic representations. In contrast to conventional zero-shot settings with pre-defined category names, we rely on a myriad of attribute combinations instead. Further, this inspired us to also sample new attributes in the model design. Hence, during training, some new attribute combinations $\bm{a}^{unseen}$ are dynamically sampled in each iteration. The sampling for the binary attributes follows a Bernoulli distribution, and a 1-trial multinomial distribution for the multi-valued attributes (i.e., mutually-exclusive attributes) according to the training data probability. Once the attribute feature extraction is complete, $f_a$ in the objective function is replaced with $[f_a,f_{a^{unseen}}]$, $f_{a^{unseen}}$ is used to generate synthetic visual feature via $G_{V}$. The synthetic feature pairs are then used for the final common space cross-modal learning.
\section{Experiments}
\begin{figure}[t]
	\centering
	\includegraphics[trim={0 11cm 0 0},clip,page=3,width=\textwidth]{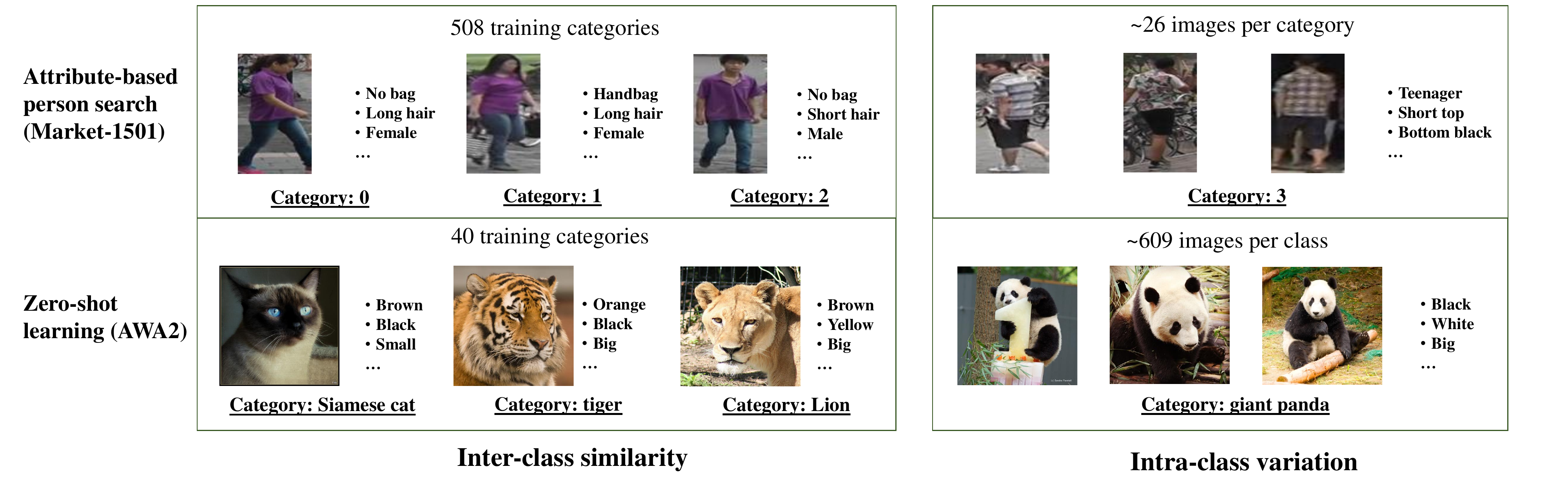}
	\caption{\footnotesize
	Example images from the person search Market-1501~\cite{zheng2015scalable} dataset and compared with the general zero-shot AWA2~\cite{xian2018zero} dataset.
	 }
	\label{fig:dataset}
\end{figure}

%
\noindent {\bf Datasets.} To evaluate SAL with an attribute-based person search task, we used two widely
adopted pedestrian attribute datasets\footnote{The DukeMTMC dataset is not publicly available. }:
(1) The \textbf{\em Market-1501 attribute} dataset~\cite{zheng2015scalable} which the Market-1501 dataset annotated with $27$ attributes for each identity ~\cite{lin2017improving} - see Fig.~\ref{fig:dataset}. 
The training set contains an average of $\sim 26$ labeled images in each category. During testing, $367$ out of $529$  \textbf{($ \textbf{ 69.4\%}$)} were from \textbf{unseen} categories. 
(2) The \textbf{\em PETA} dataset~\cite{deng2014pedestrian} consists of $19,000$ pedestrian images collected from $10$ small-scale datasets of people. Each image is annotated with 65 attributes. 
Following \cite{yin2018adversarial}, we labeled the images with categories according to their attributes, then split the dataset into a training set and a gallery set by category. In the training set, there was on average $\sim 7$ labeled images in each category, and all $200$ categories in the gallery set ($\textbf{100\%}$) were \textbf{unseen}.

\noindent {\bf Performance Metric.}
The two metrics used to evaluate performance were the cumulative matching characteristic (CMC) and mean Average Precision (mAP). We followed~\cite{paisitkriangkrai2015learning} and computed the CMC on each rank k as the probe cumulative percentage of truth matches appearing at ranks $\leq \!k$. mAP measures the recall of multiple truth matches and was computed by first computing the area under the Precision-Recall curve for each probe, then calculating the mAP over all probes.

\begin{table*}[t]
	\centering
	\caption{\footnotesize Attribute-based person search performance evaluation. Best results are shown in {\bf bold}. The second-best results are \underline{underlined}. }
	\label{tab:result}
\begin{tabular}{|ll|c|c|c|c|c|c|c|c|}
\hline
{Metric (\%)}& & \multicolumn{4}{c|}{Market-1501 Attributes} & \multicolumn{4}{c|}{PETA} \\ \hline
    \multicolumn{1}{|l|}{Model}    & Reference &mAP  &rank1 &rank5 &rank10 &  mAP   & rank1 &  rank5 & rank10  \\ \hline
\multicolumn{1}{|l|}{DeepCCA \cite{andrew2013deep}}& \footnotesize ICML'13   &17.5   &30.0  &50.7   &58.1   &11.5  &14.4   &20.8    &26.3   \\ \hline
\multicolumn{1}{|l|}{DeepMAR \cite{li2015multi}} & \footnotesize ACPR'15     &8.9  &13.1   &24.9   &32.9     &12.7   &17.8   &25.6    &31.1    \\ \hline
\multicolumn{1}{|l|}{DeepCCAE \cite{wang2015deep}} & \footnotesize ICML'15  &9.7    &8.1   &24.0   &34.6      &14.5      &14.2   &22.1   &30.0  \\ \hline
\multicolumn{1}{|l|}{2WayNet \cite{eisenschtat2017linking}}& \footnotesize CVPR'17 &7.8   &11.3   &24.4   &31.5   &15.4     &23.7   &38.5    &41.9   \\ \hline
\multicolumn{1}{|l|}{CMCE \cite{li2017identity}}& \footnotesize ICCV'17 &22.8   &35.0   &51.0   &56.5    &26.2    &31.7   &39.2   &48.4    \\ \hline
\multicolumn{1}{|l|}{ReViSE \cite{tsai2017learning}}& \footnotesize ICCV'17 & 17.7   &  24.2   &  45.2   &57.6      &  31.1    &  30.5   &    \underline{57.0}  &  61.5    \\ \hline
\multicolumn{1}{|l|}{MMCC \cite{felix2018multi}}& \footnotesize ECCV'18 &  22.2   &  34.9   &  \underline{58.7}   &   \underline{70.2}   &  \underline{33.9}   &  33.5   &  \underline{57.0}   &   \underline{69.0}  \\ \hline
\multicolumn{1}{|l|}{AAIPR \cite{yin2018adversarial}}& \footnotesize IJCAI'18 &20.7  &40.3   &49.2   & 58.6  & 27.9   & \underline{39.0}   &53.6   &62.2    \\ \hline
\multicolumn{1}{|l|}{AIHM \cite{dong2019person}} & \footnotesize ICCV'19    &  \underline{24.3}    &  \underline{43.3}   &  56.7   &  64.5   &  -   &  -   &  -   &   -    \\  \hline \hline
\multicolumn{2}{|l|}{{\bf SAL} (Ours)} & \bf 29.8   &  \bf 49.0   &  \bf 68.6   &  \bf 77.5   & \bf 41.2   &  \bf 47.0   & \bf 66.5    &  \bf 74.0   \\ \hline
\end{tabular}
\label{tab:comp_arts}
\end{table*}
\noindent {\bf Implementation Details.}
SAL was implemented based on Torchreid~\cite{torchreid, zhou2019osnet, zhou2019learning} in the Pytorch framework \cite{paszke2017automatic} with ResNet-50~\cite{he2016deep} as the image feature extractor. We used fully connected (FC) layers to form the attribute feature extractor ($64$, $128$, $256$, $512$), the middle-level generators ($256$, $128$, $256$, $512$), and the encoders ($512$, $256$, $128$). Batch normalization and a ReLU nonlinear activation followed each of the first three layers with a Tanh (activation) before the output. The discriminators were also FC layers ($512$, $256$, $1$) with batch normalization and a leaky ReLU activation following each of the first two layers, and a Sigmoid activation prior to the output. After the common space embedding, the classifiers output with 1-FC layer. 
{\bf Training process:}
 We first pre-trained the image branch and \textit{Embed} model as a person search baseline (Fig.~\ref{sec:Embedding}) until it converges. Then we trained the full SAL model for $60$ epochs with a learning rate of $0.001$ for the image branch and $0.01$ for the attribute branch.
 We chose Adam as the training optimizer~\cite{kingma2014adam} with a batch size of $128$. During testing, we calculated the cosine similarity between the query attributes and the gallery images in the common embedding space with $128$-D deep feature representations for matching.  
{\bf Training time:} It took 77 minutes for SAL to converge ($26.3$M parameters) compared to 70 minutes for a single adversarial model Embed+adv to converge ($25.0$M parameters). Both training processes were run on the same platform with $2$ NVIDIA 1080Ti GPUs. 

\subsection{Comparisons to the State-Of-The-Arts}
The results of the search task appear in Table~\ref{tab:comp_arts}, showing SAL’s performance in comparison to $9$ state-of-the-art models across $4$ types of methods. These are: 
(I) {\it attribute recognition} based method: (1) DeepMAR~\cite{li2015multi}.
(II) {\it correlation} based methods: (2) Deep Canonical Correlation Analysis (DCCA)~\cite{andrew2013deep}; (3) Deep Canonically Correlated Autoencoders (DCCAE)~\cite{wang2015deep}; (4) 2WayNet~\cite{eisenschtat2017linking}
(III) {\it common space embedding} : (5) Cross-modality Cross-entropy (CMCE)~\cite{li2017identity}; (6) ReViSE~\cite{tsai2017learning}; (7) Attribute-Image Hierarchical Matching (AIHM)~\cite{dong2019person};
(IV) {\it adversarial learning}: (8) Multi-modal Cycle-consistent (MMCC) model~\cite{felix2018multi}; (9) Adversarial Attribute-Image Person Re-ID (AAIPR)~\cite{yin2018adversarial}.

From the results, we made the following observations: (1) SAL outperformed all $9$ state-of-the-art models on both datasets in terms of mAP. On the Market-1501 dataset, the improvement over the second best scores were $5.5\%$ ($29.8$ v ${24.3}$) and $5.7\%$  (${49.0}$ v ${43.3}$) for rank1. On the PETA dataset, the improvement was $7.3\%$ (${41.2}$ v ${33.9}$) and $8.0\%$ (${47.0}$ v ${39.0}$) for rank1. This illustrates SAL’s overall performance advantages in cross-modal matching for attribute-based person search.
(2) Directly predicting the attributes using the existing recognition models and matching the predictions with the queries is not efficient (e.g., DeepMAR). This may be due to the relatively low prediction dimensions and the sparsity problems with attributes in semantic space. 
(3) Compared to the common space learning-based method (CMCE), the conventional correlation methods (e.g., DCCA, DCCAE, 2WayNet) witnessed relatively poor results. This demonstrates the power of common space learning with a common embedding space. It is worth mentioning that CMCE is specifically designed to search for people using natural language descriptions, yet SAL outperformed CMCE by $7.0\%$ mAP/$14.0\%$ rank1 with Market-1501, and by $15.0\%$ mAP/$15.3\%$ rank1 with PETA. 
(4) The adversarial model comparisons, AAIPR and MMCC, did not fare well against SAL, especially AAIPR, which utilizes single adversarial learning to align the common space. This directly demonstrates the advantages of our approach with symbiotic adversarial design compared to traditional adversarial learning methods. 
(5) Among the compared state-of-the-arts, MMCC, ReViSE and AIHM addressed {\it zero-shot problems}. Against MMCC, SAL outperformed by $7.6\%$ mAP/$14.1\%$ rank1 on Market-1501 and $7.3\%$ mAP/$13.5\%$ rank1 on PETA. AIHM is the most recent state-of-the-art in this category of methods and SAL’s performance improvement was $5.5\%$ mAP/$5.7\%$ rank1 on Market-1501. This demonstrates the advantages of SAL’s new regularization term, the common space granularity-consistency loss, for generating middle-level features.

\begin{table}[t]
    \centering
    \caption{\footnotesize Component analysis of SAL on PETA dataset.}
    \label{tab:ablation}
\begin{tabular}{|l|c|c|c|c|}
\hline
Metric (\%) & mAP & rank1 & rank5 & rank10 \\ \hline
\textit{Embed} & 31.3 & 34.0 & 57.0 & 64.5 \\ \hline
\textit{Embed} + \footnotesize\textit{adv} & 35.0 & 37.5 & 60.5 & 66.5 \\ \hline
\textit{Embed} + \footnotesize\textit{symb-adv} & 40.6 & 44.0 & 64.0 & 70.5   \\ \hline
\textit{Embed} + \footnotesize\textit{symb-adv} + \textit{unseen}\footnotesize (SAL) & 41.2 & 47.0 & 66.5 & 74.0   \\ \hline
\end{tabular}
\end{table}

\subsection{Further Analysis and Discussions}
\label{sec:model_com}

To further evaluate the different components in the model, we conduct studies on the PETA dataset.

\paragraph{\bf Component analysis of SAL}
Here, we compared: (1) the base embedding model (\textit{Embed}, Fig.~\ref{fig:retrieval}(a)), which comprises the multi-modal common space embedding base (Sec.~\ref{sec:Embedding}) and is optimized by embedding loss (Eq.~\eqref{eq:concept}) only; (2) the base embedding model plus single adversarial learning (\textit{Embed}+\textit{adv}, Fig.~\ref{fig:retrieval}(b)), 
(3) the base embedding model plus our symbiotic adversarial learning (\textit{Embed}+\textit{symb-adv}, Fig.~\ref{fig:retrieval}(c)), and 
(4) our full SAL model, which includes the attribute sampling for unseen classes. The results are shown in Table~\ref{tab:ablation}.

Compared to the \textit{Embed} model, \textit{Embed}+\textit{adv} saw an improvement of $3.7\%$ mAP/$3.5\%$ rank1, whereas \textit{Embed}+\textit{symb-adv} achieved an improvement of $9.3\%$ mAP/$10\%$ rank1, and finally SAL (\textit{Embed}+\textit{symb-adv}+\textit{unseen}) witnessed a significant improvement of $9.9\%$ mAP/$13.0\%$ rank1. 
This is a clear demonstration of the benefits of jointly considering middle-level feature augmentation and high-level cross-modal alignment instead of only having common space cross-modal alignment ($41.2\%$ vs $35.0\%$ mAP and $47.0\%$ vs $37.5\%$ rank1). We visualize the retrieved results in the supplementary material.

%

\begin{table}[t]
    \centering
    \caption{\footnotesize Effect of interactions between two GANs on PETA dataset.}
    \label{tab:ablation-loss}
\begin{tabular}{|l|c|c|c|c|}
\hline
Metric (\%) & mAP & rank1 & rank5 & rank10 \\ \hline
SAL - $L_{aug}$  & 35.4 & 38.0 & 60.0 & 69.0 \\ \hline
SAL - $L_{consis}$ & 35.2 & 39.5 & 56.5 & 66.0   \\ \hline
SAL (Full interaction) & 41.2 & 47.0 & 66.5 & 74.0   \\ \hline
\end{tabular}
\end{table}


\paragraph{\bf Effect of interactions between two GANs}
Our next test was designed to assess the influence of the symbiotic interaction, i.e., where the two GANs iteratively regularize and augment each other. Hence, we removed the interaction loss between the two GANs, which is the total augmented loss from Eq.~\eqref{eq:augtotal} and the common space granularity-consistency loss from Eq.~\eqref{eq:conc}. Removing the augmented loss (SAL - $L_{aug}$) means the model no longer uses the augmented data. Removing the common space granularity-consistency loss (SAL - $L_{consis}$) means the middle-level generators are not conditioned on high-level common embedding information, which should reduce the augmented data quality. The results in Table~\ref{tab:ablation-loss} show that, without augmented data, SAL’s mAP decreased by $5.8\%$ and by $6.0\%$ without the granularity-consistency loss as a regularizer.

\begin{table}[t]
    \centering
    \caption{\footnotesize Comparing stage-wise training vs. symbiotic training scheme.}
    \label{tab:ablation-scheme}
\begin{tabular}{|l|c|c|c|c|}
\hline
Metric (\%) & mAP & rank1 & rank5 & rank10 \\ \hline
SAL w/ stage-wise training & 35.0 & 41.0 & 58.0 & 65.0   \\ \hline
SAL w/ symbiotic training & 41.2 & 47.0 & 66.5 & 74.0   \\ \hline
\end{tabular}
\end{table}

\paragraph{\bf Comparing stage-wise training vs. symbiotic training scheme }
We wanted to gain a better understanding of the power of the symbiotic training scheme (Sec.~\ref{sec:scheme}). So, we replaced the iterative symbiotic training scheme (SAL w/ symbiotic training) with a stage-wise training (SAL w/ stage-wise training). The stage-wise training \cite{barshan2015stage} breaks down the learning process into sub-tasks that are completed stage-by-stage. So during implementation, we first train the synthesis-GAN conditioned on the common embedding. Then for the next stage, we optimise the alignment-GAN on the common space with synthetic data.
As shown in Table~\ref{tab:ablation-scheme}, there was a $6.2\%$ drop in mAP using the stage-wise training, which endorses the merit of the symbiotic training scheme. During the symbiotic training, the synthesized augmentation data and the common space alignment iteratively boost each other’s learning. A better common space constrains the data synthesis to generate better synthetic data. And, with better synthetic data for augmentation, a better common space can be learned.

\section{Conclusion}
In this work, we presented a novel symbiotic adversarial learning model, called SAL. SAL is an end-to-end framework for attribute-based person search. Two GANs sit at the base of the framework: one synthesis-GAN uses semantic representations to generate synthetic features for visual representations, and vice versa, while the other alignment-GAN optimizes the embeddings and performs cross-modal alignment on the common embedding space. Benefiting from the symbiotic adversarial structure, SAL is able to preserve finely-grained discriminative information and generate more reliable synthetic features to optimize the common embedding space. With the ability to synthesize visual representations of unseen classes, SAL is more robust to zero-shot retrieval scenarios, which are relatively common in real-world person search with diverse attribute descriptions. Extensive ablation studies illustrate the insights of our model designs. Further, we demonstrate the performance advantages of SAL over a wide range of state-of-the-art methods on two challenging benchmarks.

\paragraph{\bf Acknowledgement} This research was supported by ARC FL-170100117, IH-180100002, LE-200100049.

\clearpage
%
%
\bibliographystyle{splncs04}
\bibliography{egbib}

\newpage
\section*{Supplementary material}

 \begin{figure}[h]
   \centering
   \includegraphics[page=1,width=\textwidth]{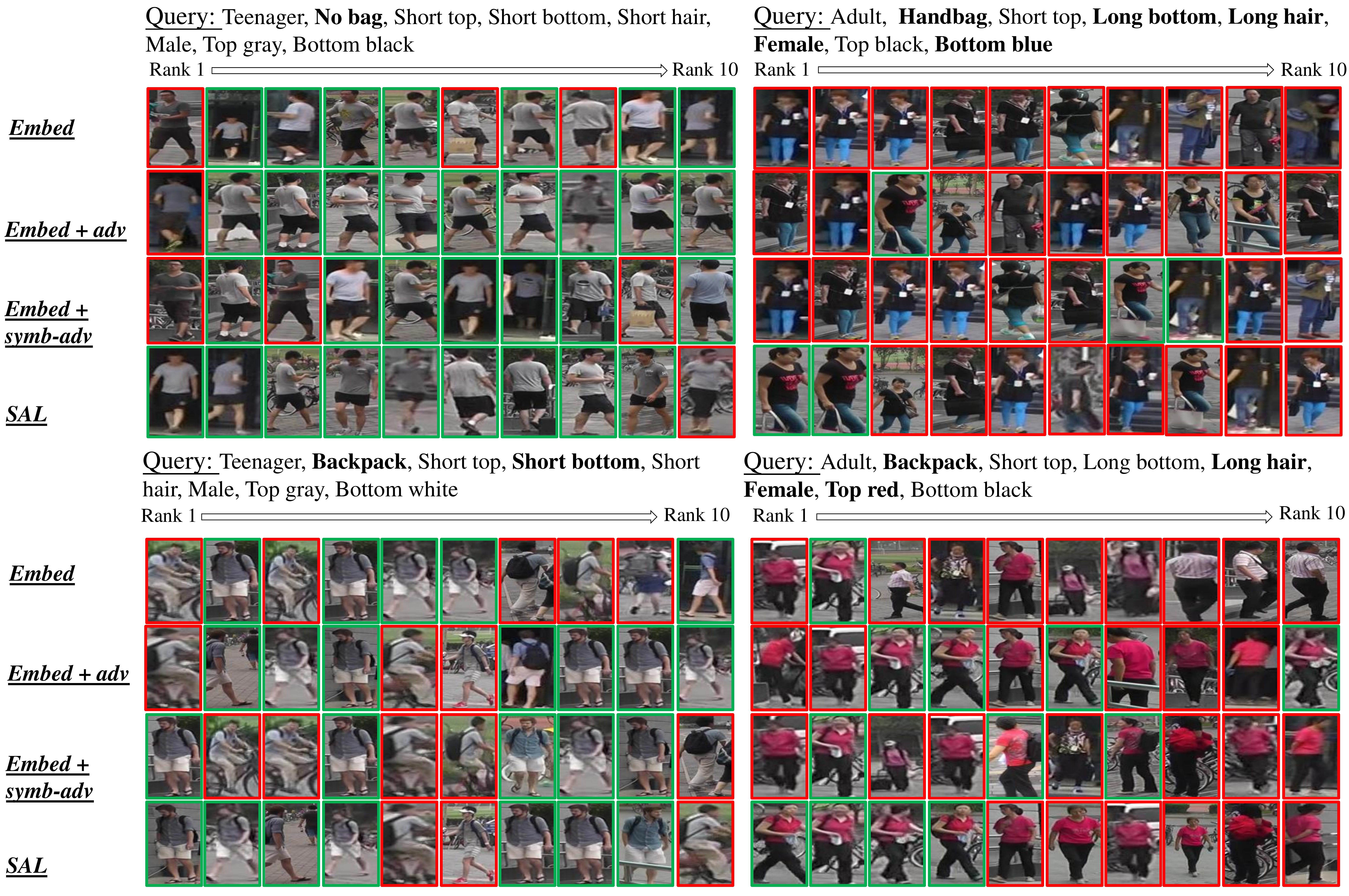}
   \caption{ \footnotesize Ranked retreival results. The query attributes are shown above the retrieved images. The \textcolor{Green}{green}/\textcolor{Red}{red} border represents correct/wrong selections respectively. The attributes in {\bf bold} correspond to false matches.  }
   \label{fig:results}
 \end{figure}
\noindent Fig.~\ref{fig:results} visualizes the ranked results from \textit{Embed}, \textit{Embed+adv}, \textit{Embed+symb-adv} and SAL to qualitatively illustrate the performance of the three models. Although the models are able to pick out some correct images from candidates in the top 10 ranks, SAL selected more with higher ranks. Checking the wrong attributes of the false ranks, we can see that SAL was better able to discern fine-grained attributes, such as different kinds of bags. Moreover, SAL picked correct images of diverse appearances, which may indicate it has some ability to overcome intra-class variation problems.

\clearpage
%
%

\end{document}